\documentclass[conference]{IEEEtran}
\IEEEoverridecommandlockouts
\usepackage{cite}
\usepackage{amsmath,amssymb,amsfonts}
\usepackage{algorithmic}
\usepackage{graphicx}
\usepackage{textcomp}
\usepackage{multicol}
\usepackage{booktabs} % For formal tables

\usepackage{lipsum}
\usepackage{xcolor}
\def\BibTeX{{\rm B\kern-.05em{\sc i\kern-.025em b}\kern-.08em
    T\kern-.1667em\lower.7ex\hbox{E}\kern-.125emX}}
\begin{document}

\title{A Multi-variable Stacked Long-Short Term Memory Network for Wind Speed Forecasting}

\author{\IEEEauthorblockN{Sisheng Liang, Long Nguyen, Fang Jin }
\IEEEauthorblockA{\textit{Department of Computer Science} \\
\textit{Texas Tech University}\\
Lubbock, Texas, USA \\
(sisheng.liang, long.nguyen, fang.jin)@ttu.edu}
}

\maketitle

\begin{abstract}

Precisely forecasting wind speed is essential for wind power producers and grid operators. However, this task is challenging due to the stochasticity of wind speed. To accurately predict short-term wind speed under uncertainties, this paper proposed a multi-variable stacked LSTMs model (MSLSTM). The proposed method utilizes multiple historical meteorological variables, such as wind speed, temperature, humidity, pressure, dew point and solar radiation to accurately predict wind speeds. The prediction performance is extensively assessed using real data collected in West Texas, USA. The experimental results show that the proposed MSLSTM can preferably capture and learn uncertainties while output competitive performance.   
\end{abstract}


\section{Introduction}
Wind energy, as one of the most promising renewable energy, has gained increasing attention in recent years. By the end of 2017, wind power capacity in operation accounts for 5.6 percent of the total electricity generation in the world according to~\cite{sawin2018renewables}. However, the fluctuating and intermittent nature of wind speed has posed many challenges to large-scale integration of wind power into power grid~\cite{buhan2016wind,fang2017high}. To mitigate the uncertain effects brought by fluctuating wind speeds, accurate short-term wind forecasting is essential for power producers and grid operators~\cite{bhaskar2012awnn}.

\paragraph{\textbf{Related work}}Existing wind prediction methods can be roughly classified into three categories: 1) Physical modeling approach, which builds models based on computational fluid dynamics utilizing detailed information such as physics of lower atmospheric layer and terrain information at the location of the wind farm. It is usually employed to predict medium-term and long-term tasks. The main drawback is that it requires physical data each time at different locations and a large amount of computation is desired. It is usually executed through a supercomputer~\cite{soman2010review}.
 2) Statistical approach, which predicts wind speed based on historical time series data. Auto-Regressive Moving Average (ARMA) model and its evolving models are among the most popular models for short-term wind prediction~\cite{jung2014current}. However, the performance declines as the prediction time steps increase.  
 3) Artificial intelligence approach, such as Artificial Neural Networks (ANN) and Recurrent Neural Networks (RNN) are used for wind and wind power prediction~\cite{khodayar2018spatio}. It is good for predicting the intermittent nature of wind due to its non-linearity. Historical meteorological data is fed into neural networks to train the parameters of the network and then predict the wind speed. In recent years, deep learning has been growing rapidly. Literature shows that compared to shallow neural network, deep learning could explore more inherent hidden patterns from data. Therefore, deep learning methods exhibit higher accuracy for wind forecasting~\cite{wang2016deep}. Recurrent neural network (RNN) was proposed to deal with time series in~\cite{khodayar2018spatio}, and the authors proposed a deep learning method based on Long Short-term Memory(LSTM) and Convolution Neural Network (CNN) that captures spatio-temporal features in wind information. RNN is designed to deal with time sequence due to the memory units in the neurons, which can remember the historical information. However, training RNN is difficult due to the fact that RNN model may not converge. LSTM is one of the improved variations of RNN, which is a much faster RNN when dealing with time sequence data and is easier to converge~\cite{hochreiter1997long}. We employ LSTM model to develop our own multi-variable learning model for wind speed forecasting.

\begin{figure}[t]
    \centering
    \includegraphics[width=\linewidth]{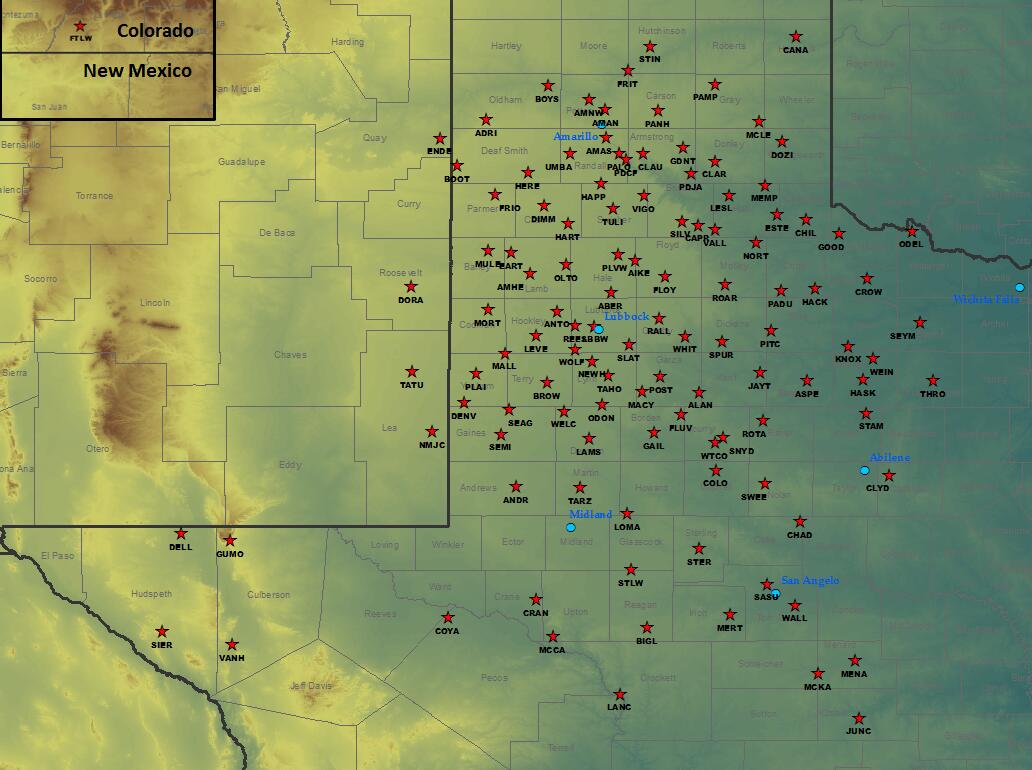}
    \caption{West Texas Mesonet Sites distribution~\cite{TexasOnline}.}
    \label{Mesonet}
\end{figure}

\begin{figure*}[t]
    \centering
    \includegraphics[width=\linewidth]{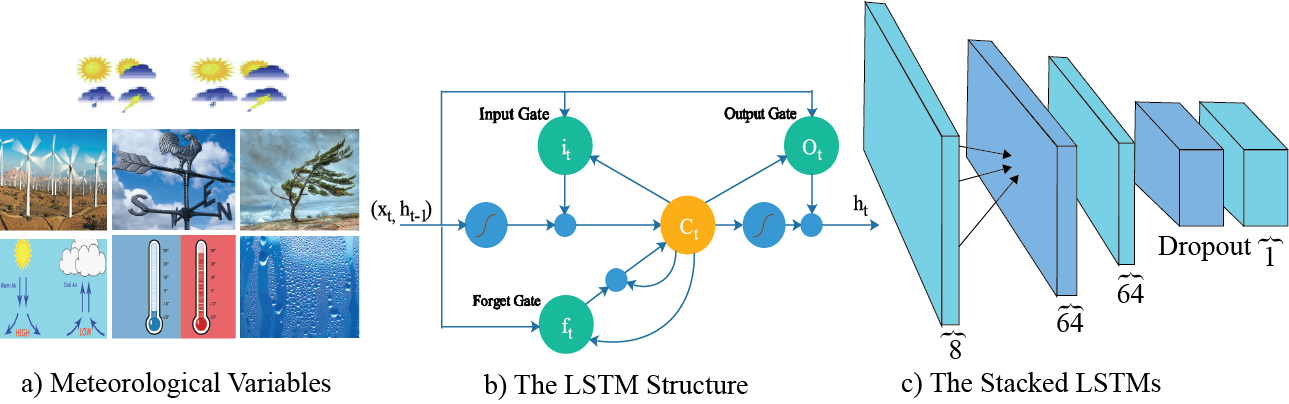}
    \caption{Visualization of the input data and used architectures. a) Meteorological variables used in this paper, including wind speed, wind direction, temperature, humidity, pressure, dew point, wind speed at 2 m and solar radiation. b) The adopted LSTM structure. c) Stacked LSTMs structure.
    % with the input layer, two LSTM layers, one dropout layer and one dense layer. 
    }
    \label{fig:architecture}
\end{figure*}

\paragraph{\textbf{Work in this paper} } A deep learning neural network was designed which takes historical wind speed, wind direction, temperature, humidity, pressure, dew point and solar radiation as inputs to predict future short-term wind speed. The reason for employing multiple meteorological variables as inputs is they could improve the accuracy of wind prediction~\cite{geerts1984short}. Even more, for some situation the prediction uncertainty depends on weather condition such as pressure~\cite{lange2001assessing}. Considering this, we incorporate multiple meteorological variables and propose a stacked long-short term memory network to learn these complex interactions for short-term wind speed forecasting. 
Extensive experiments were implemented to verified the model using real data from West Texas Mesonet Stations, as shown in Figure~\ref{Mesonet}. Different from previous work, this paper targets at wind speed forecasting at every five minutes. 
% As the best of our knowledge, this is the first work to utilize machine learning method which takes various meteorological factors into consideration for short-term wind speed prediction. 
% % No existing short-term wind prediction with machine learning method has taken all these factors as consideration.

\section{Wind Speed Forecasting Model}
%\subsection{Proposed Stacked LSTM Model}
The most challenging part of wind speed forecasting lies in its real-time dynamics, therefore, in this paper we decide to utilize LSTM model for wind speed forecasting due to its well-handling of long and short term time dependency \cite{hochreiter1997long}. 
%\subsection{LSTM Cell for Time Dependency} 
% Deep learning neural networks has been proved effective in predicting wind speed [add citation?]. Deep learning with RNN networks can better extract time patterns in dealing with time series due to the memory unit in each neuron. 
LSTM shows its superiority over the traditional recurrent neural network method. Therefore, we choose deep learning neural networks with LSTM in the hidden layer in this paper to predict wind speed. 
Figure~\ref{fig:architecture}(b) shows the basic structure of LSTM. It has an input gate $i_t$, output gate $o_t$, forget gate $f_t$ and memory cell $C_t$. Equations (1) to (5) shows how to update the output values each step\cite{8508602}. $x_t$ is the input vector and $g$ is the activation function such as ReLU or Sigmoid function. $W$ is the weight vector. We used ReLU in this paper because it will not have the problem of Vanishing Gradient and converge quickly.

\begin{equation}
    f_t = {g(W_f}{.x_t+U_f.h_{t-1}+b_f)}
\end{equation}
\vspace{-6mm}
\begin{equation}
    i_t = {g(W_i}{.x_i+U_i.h_{t-1}+b_i)}
\end{equation}
\vspace{-6mm}
\begin{equation}
    c_t = {f_t.c_{t-1}+i_t.k_t}
\end{equation}
\vspace{-6mm}
\begin{equation}
    o_t = {g(W_o.x_t+U_o.h_{t-1}+b_o)}
\end{equation}
\vspace{-6mm}
\begin{equation}
    h_t = {o_t.tanh(c_t)}.
\end{equation}

Figure \ref{fig:architecture}(c) presents the framework of the proposed MSLSTM model.
After the input layer, there are two LSTM layers stacked together before forwarding to a Dropout and a Dense layer at the final output. The first LSTM layer produces sequence vectors which will be used as the input of the subsequent LSTM layer. In addition, the LSTM layer receives feedback from its previous time step thus can capture certain patterns. In other words, this hierarchy helps the network to enable more complex representation of the wind speed time series data, and captures information at different scales. The Dropout layer excludes 5\% of the neurons to avoid over-fitting.
The proposed MSLSTM model ingests multiple meteorological variables, including Wind speed, wind direction, temperature, humidity, pressure, dew point, wind speed at two meters high and solar radiation. The deep learning neural networks could extract the hidden patterns from these variables and predict wind speed.

\begin{figure*}[t]
    \centerline{\includegraphics[width=\textwidth]{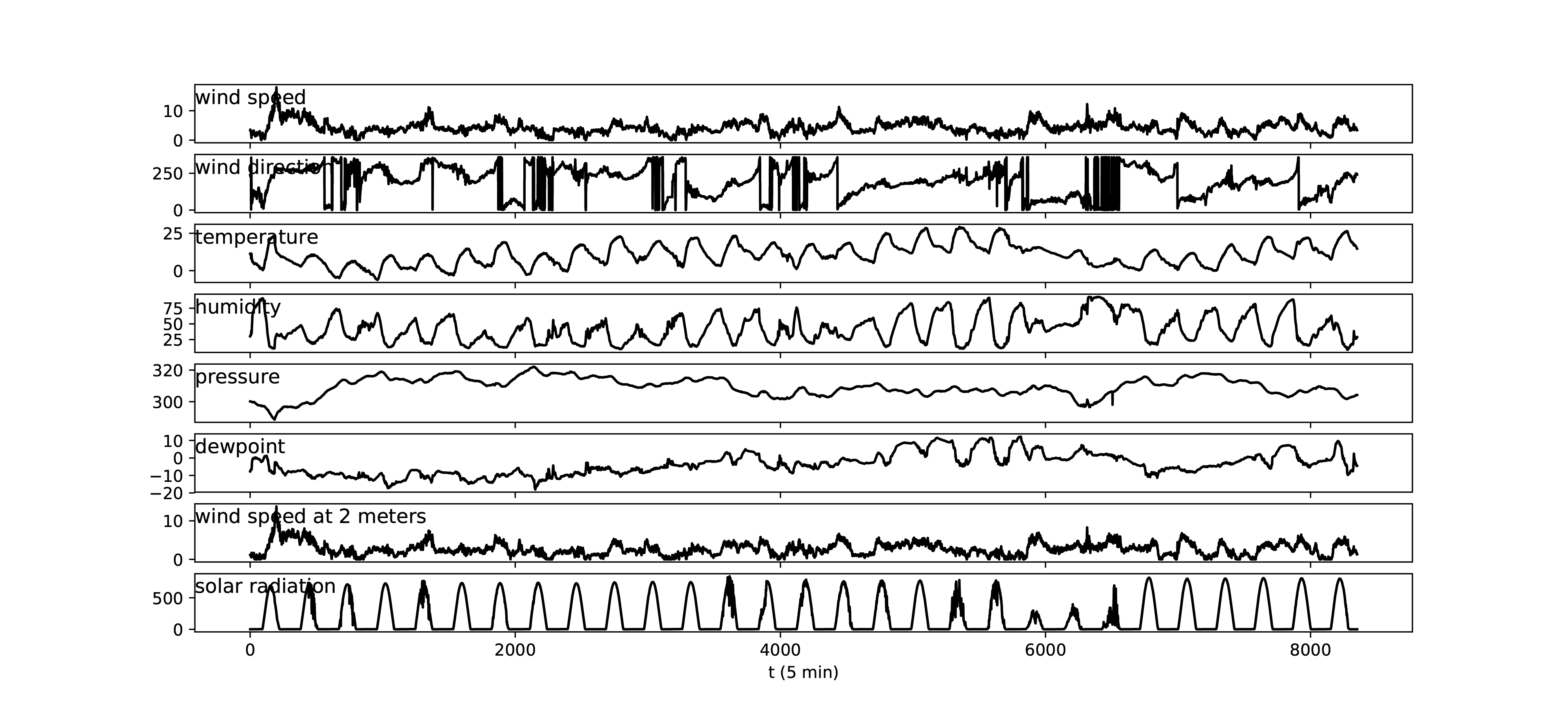}}
    \caption{Real meteorological data from one station.}
    \label{fig:weather}
\end{figure*}

\section{Experimental results}
\subsection{Dataset}
Multiple meteorological variables are used as inputs to improve the accuracy of wind prediction because there exists some correlations between different meteorological variables ~\cite{geerts1984short,lange2001assessing}.
Table~\ref{table:feature-correlation} lists the correlation between wind speed and other variables. Wind direction, temperature, solar radiation, wind speed at 2 meters shows strong correlation with wind speed (wind speed at 10 m). Other factors such as humidity and dew point also show relative high correlation with wind speed.

The dataset used in this paper was collected by West Texas Mesonet~\cite{TexasOnline}. The whole dataset includes data collected in 2016 from 117 weather stations, spreading out in West Texas. Figure~\ref{fig:weather} shows a data sample that are collected with 5-min intervals from one weather station. As shown in the table, every record in the dataset has eight weather attributes (wind speed at 10 meters, wind direction, temperature, humidity, pressure, dew point, wind speed at 2 meters, solar radiation ). %The dataset was processed and fed into the proposed deep learning network. 

\begin{table}[h]
\setlength{\tabcolsep}{16pt}
  \centering
 \caption{Correlation of variables with wind speed (* indicates significant correlations with p-value less than 2\%).}    
 \begin{tabular}{l| c | c }
    \toprule
    \toprule
          & \multicolumn{1}{ c | }{Correlation} & \multicolumn{1}{| c }{p-value} \\
  
    \midrule
    \midrule

    Wind Direction & \textbf{-0.1173*} & 0.0105  \\
    \midrule
    Temperature & \textbf{-0.1079*} & 0.0186 \\
    \midrule
     Humidity & -0.0524 & 0.2542 \\
    \midrule
    Pressure & -0.0050 & 0.9133  \\
    \midrule
    Dewpoint & -0.1021 & 0.0261 \\
    \midrule
    Wind speed at 2 meters & \textbf{-0.2719*} & 1.7064e-09 \\
    \midrule
    Solar Radiation & \textbf{0.2515*} & 2.7462e-08 \\

    \bottomrule
    \bottomrule
    \end{tabular}
  \label{table:feature-correlation}
\end{table}
\subsection{Data Preprocessing}
The data preprocessing part includes normalization and handling missing data. If there are some missing values, the corresponding records were simply deleted. The normalization is to make the input meteorological variables share the same structures and time scales. Normalized data is computed as: $ v_{ni} = (v_i-v_{min})/(v_{max}-v_{min})$, where $v_{ni}$ is normalized value for time $i$. $v_{max}$ is the maximum value of all the data for one feature. $v_i$ is real value at time $i$. $v_{min}$ is the minimum value of all the data for one feature. Wind speed, wind direction, temperature, humidity, pressure, dew point, solar radiation are all normalized to 0-1 by the above equation. 

\subsection{Implementation }
\begin{figure}[t]
    \centering
    \includegraphics[width=\linewidth]{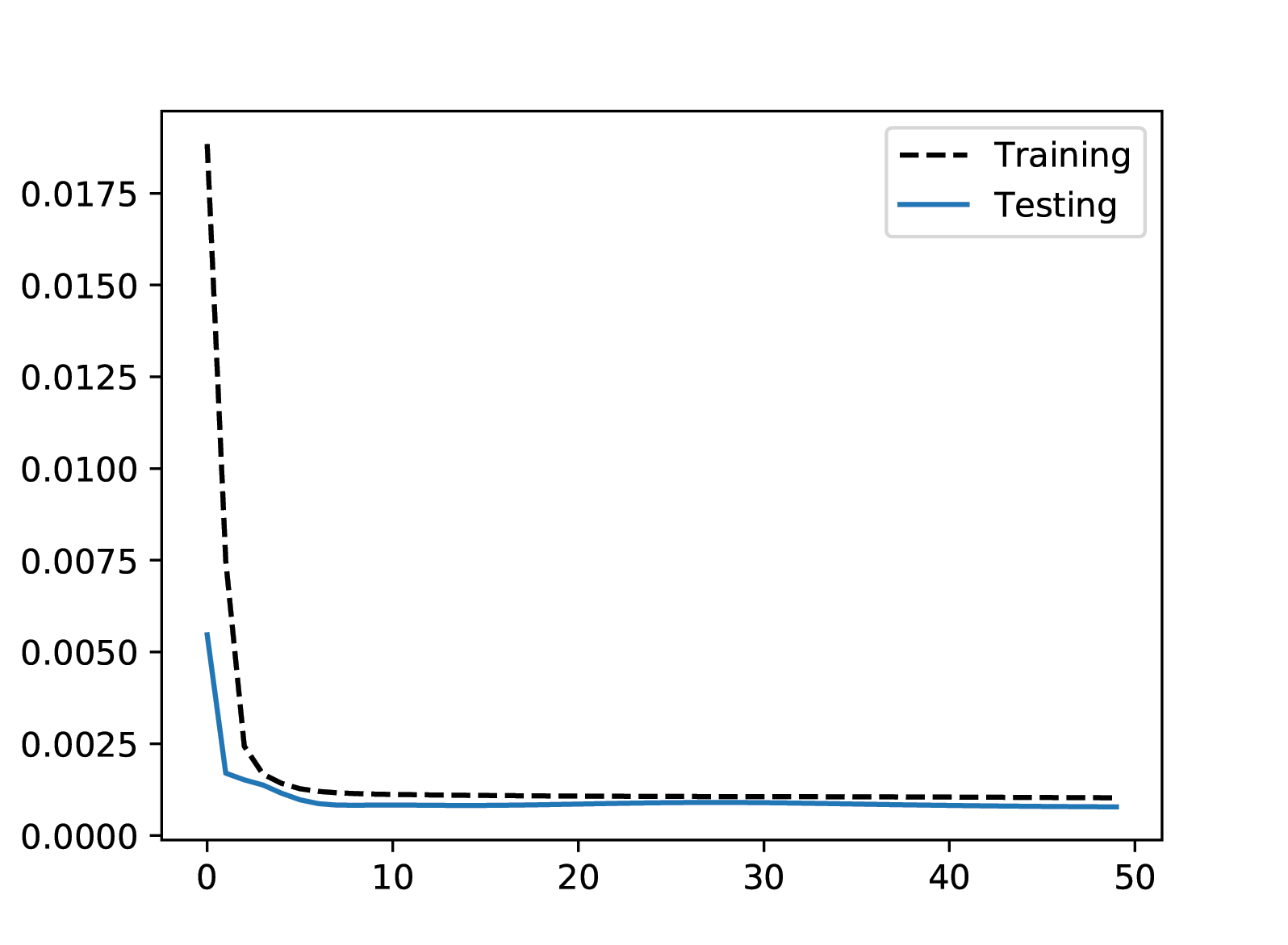}
    \caption{Lose during training and test process.}
    \label{fig:lose}
\end{figure}

The model is implemented in python with keras package which utilized Tensorflow as backend. The code is executed in a desktop with Nvidia GTX970 graphics card and i7-6700 CPU. The batch size is 40 and training epochs are 50. MSE was defined as the loss function. The loss changes of training and test process was plotted in Figure~\ref{fig:lose}. The loss of testing process is close to the loss of training process.  

\begin{figure}[h]
    \centerline{\includegraphics[width=0.5\textwidth]{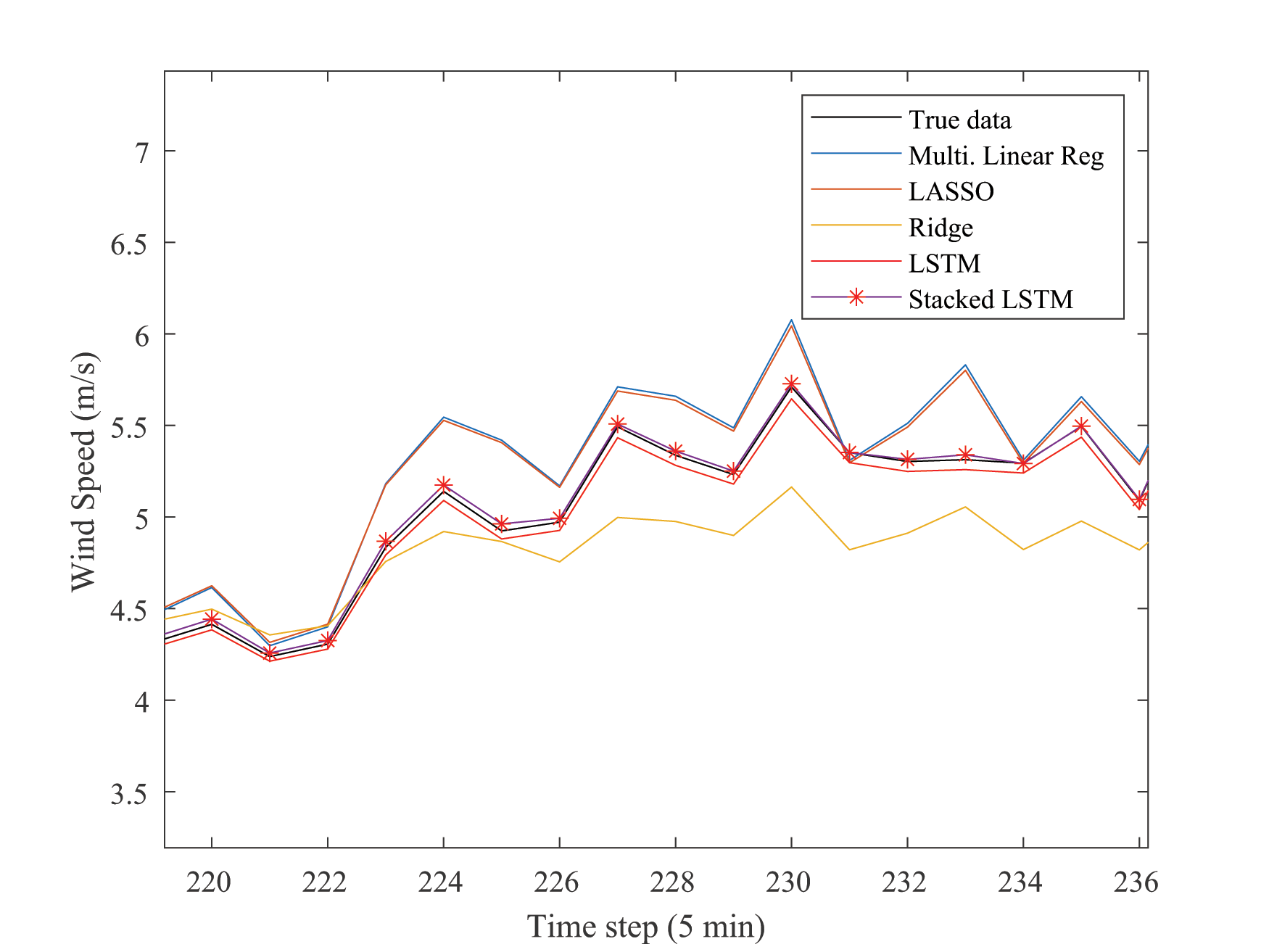}}
    \caption{Comparison of experimental results on sample data.}
    \label{result}
\end{figure}

\subsection{Evaluation Matrix}
Mean square error (MSE), root mean square error (RMSE) and mean absolute error (MAE) are employed to evaluate the performance of the models. $x_i$ denotes the real value at time t and $\widehat{x_i}$ is the predicted value at time t. $N$ is the total number of sample cycle. The MSE, RMSE and MAE values are calculated base on the following equations: $MSE =\frac{1}{N}\sum_i^N{{(x_i-\widehat{x_i})}}^2$, $RMSE =\sqrt{\frac{1}{N}\sum_i^N{{(x_i-\widehat{x_i})}}^2}$ and $MAE ={\frac{1}{N}\sum_i^N{|x_i-\widehat{x_i}|}}$.

\subsection{Numeric Results}
To verify the effectiveness of the proposed MSLSTM model, we compare the forecasting performance with multiple competing methods, including Multiple Linear Regression, LASSO, Ridge and LSTM model. The performance assessment and results shown in this paper are based on one-month data collected at Andrew in February 2016. It contains one-month data which sampled at 5-min intervals and has a total length of 8353 points. For training dataset, we randomly took 90 percent from the whole dataset, and the rest was used for testing. 

% Figure~\ref{result} illustrates prediction results from different models. Conventional prediction models such as multiple variable linear regression, LASSO, Rige are far away from true data. 
Figure~\ref{result} depicts a small piece of wind speed prediction results on sample data, where we can see conventional prediction models are far away from true data, and both LSTM and MSLSTM stick to the ground truths. Multiple Linear Regression and Lasso regression have similar results which predict higher wind speed, while Ridge regression tends to have a lower wind speed. Table~\ref{table:performance-comparison} illustrates the performance of different models under MSE, RSME and MAE matrix. We also compute $R^2$ statistics on each model, where MSLSTM has the highest value of $0.99917$ that confirms its superiority over all the baselines. 
As shown in Table~\ref{table:performance-comparison}, the evaluation results of MSE, RMSE and MAE demonstrate the stacked LSTM consistently outperforms the rest models.

% \begin{table}[htbp]
%   \centering
%   \caption{RMSE and MAE of Models for one Cycle Prediction(m/s)}
%     \begin{tabular}{llrr}
%     \toprule
%     \toprule
%     Model & \multicolumn{1}{l}{RMSE} & \multicolumn{1}{l}{MAE} \\
%     \midrule
%     LSTM& 0.484\pm{0.00005}  & 0.3591\pm{0.00001} \\
%     \midrule
%     Multi-variable LSTM & 0.4469\pm{0.00004}  & 0.3333\pm{0.00001} \\
%     \bottomrule
%     \bottomrule
%     \end{tabular}%
%   \label{tab:lossresults}%
% \end{table}%

\begin{table}[t]
\setlength{\tabcolsep}{5pt}
  \centering
 \caption{Model Performance for 5-min leading Prediction. The results are averaged from February 1 to February 29, 2016.}    
 \begin{tabular}{lr | r | r | r }
    \toprule
          & \multicolumn{1}{ c | }{MSE} & \multicolumn{1}{| c |}{RMSE} & \multicolumn{1}{| c |}{MAE} & \multicolumn{1}{| c }{$R^2$}  \\
  
    \midrule
    Multi. Linear Reg. & 0.22227 & 0.47146  & 0.37485 & 0.93750  \\
    \midrule
    Lasso & 0.20887 & 0.45703 & 0.36460 & 0.94127  \\
    \midrule
    Ridge & 0.88995 &  0.94337 & 0.73796 & 0.74979  \\
    \midrule
    LSTM & 0.01107 & 0.10522  & 0.07988 & 0.99689 \\
    \midrule
    Stacked LSTMs & \textbf{0.00297} & \textbf{0.05448} & \textbf{0.04275} & \textbf{0.99917}  \\
    \bottomrule
    \end{tabular}
  \label{table:performance-comparison}
\end{table}

\section{conclusion}
This paper proposed a multi-variable stacked LSTMs model to predict short-term wind speed. This model allows multiple meteorological parameters ingestion for real-time wind forecasting. 
The proposed MSLSTM model helps the network to enable a more complex representation of the wind speed time series data, and captures information at different scales from historical parameters, at the same time prevents over-fitting. 
All in all, the deep MSLSTM model could extract features from all the input variables which improved the precision of the prediction. MSE, RMSE, MAE and $R^2$ evaluation results indicate that MSLSTM outperformed other competing methods consistently. 
This paper uses modern machine learning techniques for short-term wind speed forecasting with extracting patterns from so many meteorological variables, and successfully learn many effective features than traditional methods, which could inspire new applications in meteorology.

\section*{Acknowledgment}
This work was supported by the U.S. National Science Foundation under Grant CNS-1737634. 
The authors would like to thank West Texas Mesonet, Dr. John Schroeder and Mr. Wes Burget for providing weather data in this paper.

\bibliographystyle{IEEEtran}
\bibliography{reference}
\vspace{12pt}
\color{red}

\end{document}